\documentclass{article}

    \PassOptionsToPackage{numbers, compress}{natbib}
 \usepackage[preprint]{neurips_2026}

\usepackage[utf8]{inputenc} % allow utf-8 input
\usepackage[T1]{fontenc}    % use 8-bit T1 fonts
\usepackage{hyperref}       % hyperlinks
\usepackage{url}            % simple URL typesetting
\usepackage{booktabs}       % professional-quality tables
\usepackage{amsfonts}       % blackboard math symbols
\usepackage{nicefrac}       % compact symbols for 1/2, etc.
\usepackage{microtype}      % microtypography
\usepackage{xcolor}         % colors

\usepackage{amsmath}
\usepackage{amssymb}
\usepackage{mathtools}
\usepackage{amsthm}
\usepackage{amsfonts}
\usepackage{xcolor}
\usepackage{booktabs}
\usepackage{multirow}
\usepackage{bbm}
\usepackage{tikz}
\usetikzlibrary{positioning, arrows.meta, fit, calc}
\usepackage{graphicx}
\usepackage{algorithm}
\usepackage{algorithmic}
\usepackage{subcaption}
\usepackage{amssymb}
\usepackage{pifont}
\usepackage{float}

\usepackage[most]{tcolorbox}
\usepackage{xcolor}
\usepackage{listings}

\definecolor{promptblue}{RGB}{35,72,120}
\definecolor{promptgray}{RGB}{70,70,70}
\definecolor{promptgreen}{RGB}{34,120,72}
\definecolor{promptorange}{RGB}{160,96,24}

\newtcblisting{promptbox}[2][]{%
  enhanced,
  breakable,
  listing only,
  listing engine=listings,
  colback=gray!3,
  colframe=#1,
  colbacktitle=#1,
  coltitle=white,
  fonttitle=\bfseries,
  title={#2},
  boxrule=0.8pt,
  arc=2mm,
toptitle=0.8mm,
bottomtitle=0.6mm,
left=3mm,
right=3mm,
top=1.8mm,
bottom=1.5mm,
  listing options={
    basicstyle=\ttfamily\small,
    columns=fullflexible,
    keepspaces=true,
    breaklines=true,
    breakatwhitespace=false,
    showstringspaces=false
  }
}

\title{KEPO: Knowledge-Enhanced Preference Optimization for Multimodal Reasoning with Applications to Medical VQA}

\author{%
  Fan Yang \\
  Chapman University\\
  Orange, CA, USA \\
  \And
  Rui Meng \\
  Lawrence Berkeley National Laboratory \\
  Berkeley, CA, USA \\
  \AND
  Trudi Di Qi \\
  Chapman University \\
  Orange, CA, USA \\
  \And
  Ali Ezzati \\
  University of California, Irvine \\
  Irvine, CA, USA \\
  \And
  Yuxin Wen\thanks{Correspondence to: Yuxin Wen
<yuwen@chapman.edu>.} \\
  Chapman University \\
  Orange, CA, USA \\
}

\begin{document}

\maketitle

\vspace{-0.5cm}
\begin{abstract}
Reinforcement learning (RL) has emerged as a promising paradigm for inducing explicit reasoning behaviors in large multimodal vision-language models.
However, multimodal reasoning-oriented RL post-training remains fundamentally challenging due to sparse trajectory-level rewards, leading to ambiguous credit assignment and severe exploration failures that can trap the policy in a ``learning cliff.''
Recent on-policy distillation methods introduce dense teacher supervision to stabilize optimization, but apply it uniformly across all generated trajectories.
We argue that such uniform distillation is ill-suited for reasoning-intensive tasks, as low-quality on-policy trajectories often originate from early logical errors, and distillation under flawed contexts injects noisy and misaligned gradients.
To address these challenges, we propose Knowledge-Enhanced Preference Optimization (KEPO), a unified post-training framework that integrates:
(i) a quality-gated on-policy distillation objective that selectively applies dense teacher guidance only to high-quality trajectories, and
(ii) a knowledge-enhanced exploration strategy that leverages hints learned from a teacher model to rejectively sample reward-positive candidate trajectories for RL, thereby mitigating exploration collapse.
Evaluated on a challenging medical multimodal visual question answering benchmark under single-source generalization, KEPO demonstrates improved training stability, more coherent reasoning behaviors, and superior out-of-distribution performance over reinforcement learning and on-policy distillation baselines.
\end{abstract}

\vspace{-0.5cm}
\section{Introduction}

\vspace{-0.2cm}
Recent advances in large multimodal vision-language models (VLMs) have highlighted the importance of post-training techniques that induce explicit reasoning behaviors, particularly in tasks requiring multi-step inference~\cite{lai2025medr1, pan2025medvlmr1}.
Reinforcement learning (RL) has emerged as a powerful paradigm for such post-training~\cite{ouyang2022training,Guo_2025}, enabling models to explore and optimize complex decision trajectories using task-level rewards.
This capability is especially critical in high-stakes domains such as healthcare, where errors in intermediate reasoning steps may lead to significant downstream consequences.
In such settings, reliable reasoning requires not only correct final answers, but also coherent and robust intermediate decision-making processes.

Despite its success, reinforcement-based post-training for reasoning remains notably challenging.
Most existing methods rely on sparse, trajectory-level reward signals, which introduce two persistent failure modes.
First, sparse rewards lead to ambiguous credit assignment in long-context reasoning,
where a single local mistake can invalidate an otherwise coherent chain~\cite{lightman2023let}.
Second, under standard on-policy reinforcement learning, sparse supervision exacerbates
exploration failure: naive exploration often fails to expose the model to any
reward-bearing trajectories when rewards are rare or binary, trapping the policy in a cold-start regime with no informative reasoning signals~\cite{xuechen2025bread, zhang2025stephint}.
We refer to this phenomenon as a learning cliff regime in this work, which has been observed in recent studies of reinforcement learning with sparse or binary rewards~\cite{zhang2025scafgrpo, yu2025dapo}.

To alleviate the limitations of sparse rewards, recent work has explored integrating dense supervision into reinforcement learning.
In particular, on-policy knowledge distillation methods~\cite{agarwal2024gkd, bousselham2025vold}
propose leveraging a strong teacher model to provide token-level guidance on trajectories
sampled from the student policy.
While such approaches improve optimization stability, they are not designed to explicitly
model the dynamics of reasoning emergence under sparse rewards.
In reasoning-intensive settings, incorrect trajectories often arise from logical fallacies
or hallucinations early in the chain-of-thought.
Distilling these trajectories forces the student to mimic teacher distribution conditioned on a flawed context, injecting noisy and potentially conflicting gradient signals that can hinder learning.
This issue is further amplified in multimodal vision-language reasoning, where errors may stem jointly from visual perception and language generation.

In this work, we propose Knowledge-Enhanced Preference Optimization (KEPO), a unified post-training framework that addresses both optimization instability and exploration failure in reinforcement-based reasoning.
KEPO consists of two tightly coupled components.
First, we introduce a quality-gated on-policy distillation objective that selectively
applies dense teacher supervision only to reward-aligned trajectories, transforming distillation from a competing objective into a mechanism for fine-grained credit assignment.
Second, we propose a knowledge-enhanced exploration strategy that actively injects teacher-guided trajectories into the on-policy rollout process when naive exploration fails, enabling the policy to escape the learning cliff without relying on static offline supervision.
Importantly, KEPO treats distillation as a structured auxiliary signal that is integrated into reinforcement learning to guide reasoning behavior in a reward-aligned manner.

We evaluate KEPO on a challenging low-resource, cross-modality medical visual question answering setting~\cite{hu2024omnimedvqa, lai2025medr1, pan2025medvlmr1},
training on a single in-domain modality and testing across diverse out-of-distribution modalities.
Empirical results show that KEPO improves training stability, accelerates the emergence of effective reasoning behavior, and achieves superior out-of-distribution generalization compared to reinforcement learning and on-policy distillation baselines.
Overall, KEPO provides a principled framework for integrating dense teacher guidance into reinforcement-based post-training, addressing challenges posed by sparse rewards and insufficient exploration in reasoning-oriented learning.

\vspace{-0.2cm}
\paragraph{Contributions.}
In summary, our main contributions are:
\vspace{-0.2cm}
\begin{itemize}
    \item We propose Knowledge-Enhanced Preference Optimization (KEPO), a unified post-training framework that integrates quality-gated teacher supervision with adaptive exploration to enhance reinforcement learning for multimodal vision-language models.
\vspace{-0.1cm}
    \item We identify two fundamental failure modes in reasoning-oriented reinforcement learning:
    contextual noise from uniform on-policy distillation over flawed trajectories, and exploration stagnation under sparse rewards.
    We show that selectively gating distillation by trajectory quality and injecting hint-aware trajectories are key to resolving these issues.
\vspace{-0.1cm}
    \item We instantiate recent on-policy distillation frameworks~\cite{agarwal2024gkd} in the multimodal vision-language setting to enable principled comparisons, providing a unified experimental basis for studying distillation-based and reinforcement-based post-training approaches.
\vspace{-0.1cm}
    % \item We conduct extensive experiments on a challenging medical Visual Question Answering (VQA) benchmark~\cite{hu2024omnimedvqa} under a low-resource, single-source domain generalization setting, demonstrating that KEPO induces robust reasoning behaviors and achieves superior out-of-distribution generalization compared to standard reinforcement learning and on-policy distillation baselines.
    \item We conduct extensive experiments on a medical Visual Question Answering (VQA) benchmark~\cite{hu2024omnimedvqa} under a low-resource, single-source domain generalization setting, 
    showing that KEPO yields robust reasoning and superior out-of-distribution generalization over reinforcement learning and on-policy distillation baselines.
\end{itemize}

\vspace{-0.3cm}
\section{Related Work}

\vspace{-0.2cm}
\subsection{Reinforcement Learning and the Sparse Reward Challenge}
\vspace{-0.2cm}
Reinforcement learning has become a central paradigm for post-training large language models and multimodal vision-language models, particularly in aligning model behavior with task-specific objectives~\cite{ouyang2022training}. Policy-gradient methods such as Proximal Policy Optimization (PPO) and its variants have been widely adopted due to their empirical stability. In reasoning-oriented tasks, Group Relative Policy Optimization (GRPO)~\cite{shao2024deepseekmath} further improves training stability by leveraging group-wise normalization over multiple sampled trajectories.
However, these methods primarily focus on trajectory-level reward signals. Such sparse supervision introduces severe challenges in credit assignment, especially for long-context reasoning tasks where a single local error can invalidate an otherwise correct reasoning chain. 
This sparsity often results in high-variance gradient estimates and unstable optimization dynamics, motivating exploration of richer supervision signals within RL-based post-training frameworks, particularly in multimodal reasoning settings.
Most existing RL post-training methods are primarily studied in text-only settings, while multimodal reasoning poses additional challenges for credit assignment and exploration.

\vspace{-0.2cm}
\subsection{Dense Supervision via On-Policy Distillation}
\vspace{-0.2cm}
To alleviate the limitations of sparse rewards, Knowledge Distillation (KD) has been explored as a means of providing dense supervision from strong teacher models~\cite{hinton2015distilling}. While standard supervised KD suffers from exposure bias, Generalized Knowledge Distillation (GKD)~\cite{agarwal2024gkd} and subsequent studies~\cite{thinkingmachines2025, huggingface2025} propose on-policy distillation, where the student learns from trajectories sampled from its own policy.
Beyond pure distillation, hybrid paradigms such as Prefix-RFT~\cite{huang2025blending} blend supervised and reinforcement fine-tuning to combine demonstration with exploration.
In the multimodal setting, VOLD~\cite{bousselham2025vold} extends on-policy distillation to vision-language models.
Crucially, however, these methods typically apply distillation uniformly across all generated trajectories. As we argue in this work, this assumption breaks down in reasoning-intensive tasks: distilling low-quality trajectories conditioned on flawed contexts injects noisy gradients, highlighting the critical need for quality-aware mechanisms that restrict dense supervision to reward-aligned paths.

\vspace{-0.2cm}
\subsection{Structured Exploration and Credit Assignment}
\vspace{-0.2cm}
Beyond distillation, extensive research has focused on resolving credit assignment and exploration stagnation. Process Reward Models (PRMs)~\cite{lightman2023let} assign scores to intermediate steps to alleviate ambiguity but require costly annotations.
To facilitate exploration without full supervision, recent methods introduce structured guidance. For example, BREAD~\cite{xuechen2025bread} uses branched rollouts with partial demonstrations,  StepHint~\cite{zhang2025stephint} employs stepwise hints to mitigate near-miss rewards. Similarly, Scaffolded GRPO~\cite{zhang2025scafgrpo} injects adaptive guidance, and LUFFY~\cite{yan2025learning} incorporates off-policy traces from stronger teachers.
Our work complements these directions but differs in a key aspect: rather than relying on external reward models or static demonstrations, we leverage the teacher model dynamically to guide exploration and provide dense credit assignment specifically when naive exploration fails, particularly in multimodal reasoning settings with more complex exploration dynamics.

\vspace{-0.2cm}
\enlargethispage{\baselineskip}
\subsection{Reasoning in Medical Vision-Language Models}
\vspace{-0.2cm}
Medical Vision-Language Models (Med-VLMs) have emerged as a prominent application domain for evaluating multimodal reasoning capabilities~\cite{hu2024omnimedvqa}. Early approaches primarily relied on supervised fine-tuning with large-scale datasets~\cite{chen2024huatuogpt, liu2024improved}, which often struggle with multi-step reasoning and out-of-distribution generalization.
More recent efforts, such as Med-R1~\cite{lai2025medr1} and MedVLM-R1~\cite{pan2025medvlmr1}, adopt reinforcement learning (e.g., GRPO) to induce explicit reasoning traces.
However, these methods largely rely on sparse, outcome-based rewards.
Medical VQA serves as a natural and challenging testbed for
reasoning-oriented post-training, as its performance cannot
be reliably improved through domain-specific engineering alone, but instead requires learning transferable reasoning patterns that generalize across modalities.
By strictly controlling the training source to a single modality (MRI-only), we isolate cross-modality generalization as the primary evaluation axis,
enabling a focused comparison of reasoning-oriented post-training methods under sparse-reward supervision.

\vspace{-0.3cm}
\section{Method}

\begin{figure*}[t] 
    \vspace{-0.9cm} 
    \centering
\includegraphics[width=0.99\textwidth]{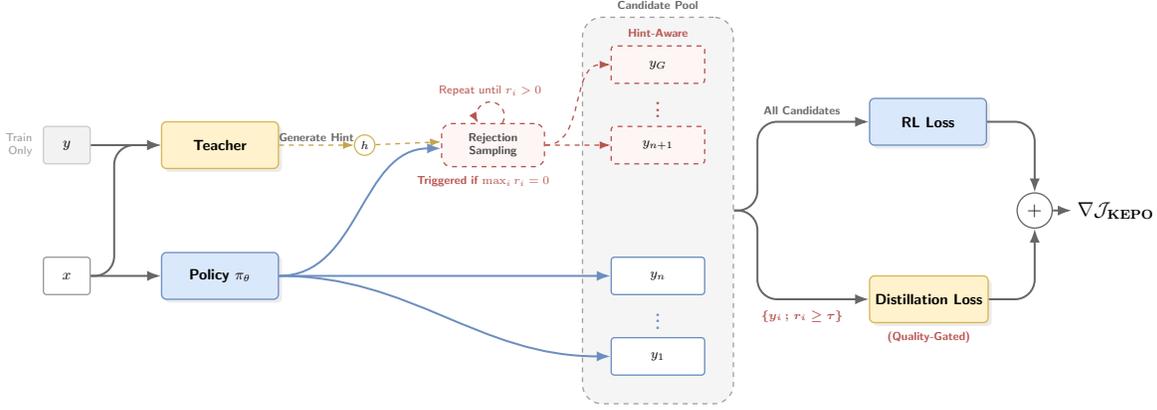} 
    
    \vspace{-0.1cm} 
    
\caption{Overview of the KEPO framework.
KEPO augments reinforcement-based post-training by jointly integrating knowledge-enhanced exploration and quality-gated distillation.
Given an input $(x, y)$ during training, the student model adaptively triggers knowledge-enhanced exploration when all sampled trajectories receive zero rewards.
During exploration, a teacher model first generates auxiliary reasoning hints. 
% Then, within the temporary recovery branch, conditioned on the hint and privileged answer context available only during training, the student model generates reward-positive candidates $(y_{n+1}, \dots, y_{G=n+m})$ via a rejection sampling procedure. 
Then, within the temporary recovery branch, the student model generates candidate trajectories conditioned on the hint and privileged answer context available only during training, while rejection sampling retains reward-positive candidates $(y_{n+1}, \dots, y_G)$, where $G=n+m$ denotes the final pool composition.
Together with the standard on-policy rollouts $(y_1, \dots, y_n)$, the candidate pool contains $G$ responses.
Finally, the reinforcement learning objective is optimized over the full pool, while an auxiliary on-policy distillation loss is selectively applied only to high-quality samples satisfying $r_i \geq \tau$.
The student policy is updated by jointly optimizing the resulting objective $\mathcal{J}_{\text{KEPO}}$.}

    \label{fig:framework}
    \vspace{-0.5cm} 
\end{figure*}

\vspace{-0.2cm} 
\subsection{Preliminaries: Group-Based Policy Optimization}
\label{sec:preliminaries}
\vspace{-0.2cm}

We consider the standard RL fine-tuning setting, where a vision-language model policy $\pi_\theta$ is optimized to maximize the expected reward of its generated outputs. The general RL objective is defined as
\begin{equation}
    \mathcal{J}_{\text{RL}}(\theta)
    :=
    \mathbb{E}_{x \sim X}\mathbb{E}_{y \sim \pi_\theta(\cdot \mid x)}[r(y)],
\end{equation}
where $x \sim X$ denotes an input prompt sampled from the training distribution, $y$ denotes a generated trajectory, $r(y)$ is a scalar reward.

To improve training stability and reduce gradient variance, recent group-based policy optimization methods sample multiple trajectories for each input prompt and perform group-wise updates. A unified Monte-Carlo formulation for such methods can be expressed as
\begin{equation}
\begin{aligned}
\mathcal{J}_{\text{MCRL}}&(\theta)
=
\mathbb{E}_{x \sim X,\; y_{1:G} \sim \pi_{\theta_{\mathrm{old}}}(\cdot \mid x)}
\Big[
\sum_{i=1}^{G}
w_i(\theta)\, \hat{A}_i 
 -\beta \,
D_{\mathrm{KL}}
\!\left(
\pi_\theta(\cdot \mid x)
\,\Vert\,
\pi_{\mathrm{ref}}(\cdot \mid x)
\right)
\Big].
\end{aligned}
\end{equation}
Here, $G$ is the number of samples for Monte Carlo rollout. $D_{\mathrm{KL}}(\cdot\Vert\cdot)$ is the Kullback--Leibler (KL) divergence, $\beta$ controls the strength of KL regularization,  and $\pi_{\mathrm{ref}}$ is a fixed reference policy.
The importance weight $w_i(\theta)$ corrects for the policy update from the previous policy $\pi_{\theta_{\mathrm{old}}}$ to the current policy $\pi_\theta$:
\begin{equation}
w_i(\theta)
=
\frac{\pi_\theta(y_i \mid x)}
{\pi_{\theta_{\mathrm{old}}}(y_i \mid x)}.
\end{equation}

Different group-based algorithms primarily differ in how the advantage term $\hat{A}_i$ is estimated.
Reinforcement Learning with Leave-One-Out (RLOO)~\cite{ahmadian2024back} adopts a leave-one-out baseline constructed from the remaining samples within the same group, yielding an unbiased and low-variance estimator without requiring a learned value function:
\begin{equation}
\hat{A}_i^{\text{RLOO}}
=
r(y_i)
-
\frac{1}{G-1}
\sum_{\substack{j=1 \\ j \neq i}}^{G} r(y_j).
\end{equation}

Alternatively, Group Relative Policy Optimization (GRPO) employs clipped importance weights to further stabilize policy updates, together with a group-mean baseline for advantage estimation:
\begin{equation}
\begin{aligned}
w_i(\theta)
&=
\min\!\Big(
r_i(\theta),
\operatorname{clip}
\big(
r_i(\theta),
1 - \epsilon,
1 + \epsilon
\big)
\Big), \\
r_i(\theta)
&=
\frac{\pi_\theta(y_i \mid x)}
{\pi_{\theta_{\mathrm{old}}}(y_i \mid x)},
\end{aligned}
\end{equation}
and the advantage is estimated by
\begin{equation}
\hat{A}_i
=
r(y_i)
-
\frac{1}{G}
\sum_{j=1}^{G} r(y_j).
\end{equation}

These group-based formulations provide a common and stable optimization backbone for preference-based policy learning, upon which our method is built.

\vspace{-0.2cm} 
\subsection{Knowledge-Enhanced Preference Optimization}
\vspace{-0.2cm}
\label{sec:kepo}
Reinforcement-based post-training commonly suffers from two failure modes in challenging reasoning regimes:
(i) sparse trajectory-level rewards, which hinder stable optimization and precise credit assignment, and
(ii) insufficient exploration, where naive on-policy sampling fails to discover reward-bearing trajectories.
To address these issues in a unified framework, the proposed Knowledge-Enhanced Preference Optimization (KEPO) consists of two tightly coupled components:
(i) a quality-gated distillation objective that stabilizes optimization through dense teacher guidance, and
(ii) a knowledge-enhanced exploration mechanism that adaptively exposes the policy to reward-bearing trajectories when naive exploration fails.
We describe these two components in detail below.

\vspace{-0.2cm} 
\subsubsection{Preference Optimization meets Quality-Gated Distillation}
\vspace{-0.2cm}
\label{sec:kepo-objective}
While group-based RL methods effectively stabilize policy updates, their learning signal is confined to trajectory-level feedback, offering no explicit token-level guidance during optimization.
In contrast, supervised fine-tuning (SFT) benefits from dense, per-token supervision but lacks the exploration capability inherent to reinforcement learning.
This contrast motivates incorporating dense teacher guidance into the RL framework as an auxiliary learning signal.
Our formulation is inspired by recent advances in on-policy distillation for language models~\cite{agarwal2024gkd}, while addressing distinct challenges arising in reinforcement-based post-training of multimodal vision-language models.
% \textcolor{red}{to be replaced to the following:
% Specifically, we formulate the KEPO objective, which leverages both sparse and dense rewards on the on-policy data at the same time.
% }
Specifically, we formulate the KEPO objective, which jointly leverages sparse trajectory rewards and dense distillation supervision on on-policy data.
% Specifically, as a core optimization component of KEPO, we formulate a knowledge-aware preference optimization objective (KAPO) that integrates on-policy distillation as an auxiliary signal during policy learning.
The key idea is to selectively apply knowledge distillation on high-quality trajectories generated by the current policy, as determined by a reward-based threshold during policy learning. Formally, our objective is defined as
% However, standard RL methods operate primarily at the trajectory level, leading to sparse rewards. First, it could cause severe credit assignment ambiguity, in particular to long context settings, where only a few actions matter but the credits are uniformly distributed for all tokens. Second, it could amplify variance of the gradient estimator, leading to unstable training. Thus, it is necessary to take benefits from the dense per-token level signals akin to SFT. To take the benefits from both worlds, we incorporate the dense per-tokens level rewards into the RL algorithms by leveraging the on policy distillation objective. 
% \vspace{-0.3cm} 
\begin{equation}
\begin{aligned}
\mathcal{J}_{\text{KEPO}}&(\theta)
=
\mathbb{E}_{x \sim X,\; y_{1:G} \sim \pi_{\theta_{\mathrm{old}}}(\cdot \mid x)}
\Bigg[
\sum_{i=1}^{G}
w_i(\theta)\, \hat{A}_i 
 -\;
\mathbb{I}_{r_i \geq \tau}\,
\mathcal{D}
\!\left(
\pi_T \,\Vert\, \pi_\theta
\right)
(y_i \mid x) \\
& \quad -\;
\beta\,
D_{\mathrm{KL}}
\!\left(
\pi_\theta(\cdot \mid x)
\,\Vert\,
\pi_{\mathrm{ref}}(\cdot \mid x)
\right)
\Bigg],
\end{aligned}
\end{equation}
where $\pi_T$ denotes a fixed teacher model and $\tau$ specifies the quality threshold for activating the distillation term. 
% We expect that a positive threshold can improve the distillation performance.
A positive threshold therefore activates quality-gated distillation.
% In our binary-reward setting, $\tau$ serves as a task-driven quality filter: $\tau=0$ recovers ungated distillation, whereas $\tau=1$ restricts distillation to reward-positive trajectories.
In our discrete-reward setting, $\tau$ serves as a task-driven quality filter: $\tau=0$ recovers ungated distillation, whereas $\tau=1$ restricts distillation to trajectories with reward $r \geq 1$ under the task-specific scoring function.
% In our discrete-reward setting, $\tau$ serves as a task-driven quality filter: $\tau=0$ recovers ungated distillation, whereas $\tau=1$ restricts distillation to trajectories with non-zero reward under the task-specific scoring function.

The proposed objective consists of three components:
(i) an advantage-weighted preference optimization term, which drives policy improvement based on relative trajectory quality,
(ii) a quality-gated on-policy distillation term that provides dense teacher guidance on reward-aligned trajectories, and
(iii) a KL regularization term that constrains policy updates to prevent excessive deviation from reference policy.
Importantly, the distillation term can be viewed as delivering dense supervision aligned with reward-consistent trajectories in tandem with the RL objective, rather than performing reward shaping by altering the reward function itself.
% Among these components, the quality-gated distillation term plays a central role by injecting dense supervision on reward-aligned trajectories, complementing sparse rewards with token-level guidance.
% In this way, the proposed objective preserves the exploration benefits of reinforcement learning while leveraging dense supervision to improve optimization stability and credit assignment.

In practical settings, particularly within specialized expert domains, the teacher model is not perfect and may itself generate flawed reasoning trajectories.
Thus, we apply distillation only to 
trajectories of sufficient quality. 
This design is motivated by two complementary considerations.
First, restricting distillation to reward-positive samples ensures that dense teacher guidance is aligned with the reinforcement learning objective, preventing the distillation signal from overriding reward-driven optimization.
Under this design, distillation acts as a form of dense credit assignment that amplifies successful behaviors rather than competing with the global reward.
Second, in reasoning-intensive tasks with chain-of-thought generation, low-quality trajectories often arise from logical errors or hallucinations early in the reasoning process, where subsequent tokens are then conditioned on an incorrect intermediate state, making it difficult for the teacher model to provide a clear learning signal.
Applying on-policy distillation to such trajectories forces the student to mimic the teacher’s distribution conditioned on a flawed context, injecting noisy and potentially conflicting gradient signals into policy learning.
This observation is consistent with recent findings that teacher supervision in on-policy distillation is not uniformly reliable across trajectories, particularly when conditioned on imperfect intermediate reasoning states~\cite{zheng2026scopesignalcalibratedonpolicydistillation}.
Accordingly, distillation is activated only for trajectories satisfying $r_i \geq \tau$ through the indicator function $\mathbb{I}_{r_i \geq \tau}$.
Together, this quality-gated design complements trajectory-level rewards with token-level guidance while avoiding supervision under flawed contexts.

\vspace{-0.2cm}
\subsubsection{Knowledge-Enhanced Exploration via Hint-Aware Rejection Sampling}
\vspace{-0.2cm}
\label{sec:kepo-exploration}
While the quality-gated distillation objective in
Section~\ref{sec:kepo-objective} effectively stabilizes optimization and improves credit assignment, KEPO pairs it with a complementary exploration mechanism to ensure exposure to reward-bearing trajectories.
In simple settings, such trajectories are readily available.
However, in complex multimodal reasoning or multi-choice scenarios, early-stage policies may seldom generate reward-bearing trajectories, resulting in sparse learning signals.
We refer to this phenomenon as a learning cliff regime, where difficult problems contribute only weakly to gradient updates due to the lack of informative exploration signals, consistent with observations in reinforcement learning with sparse or binary rewards~\cite{zhang2025scafgrpo}.

A common mitigation strategy is to perform Supervised Fine-Tuning (SFT) prior to reinforcement-based alignment.
However, recent studies suggest that the interaction between SFT and Reinforcement Fine-Tuning (RFT) is highly sensitive to training protocols and task domains, leaving principled approaches for combining offline supervision with online reinforcement learning poorly understood~\cite{cai2025backtracking, chen2025sftrl, chen2025stepwise}.
Several works explore leveraging offline datasets to assist post-training~\cite{yan2025learning, liu2025uft, ma2025interleaved, huang2025blending}, yet these methods typically rely on static data and do not directly address exploration failure during on-policy rollout.

Instead of incorporating offline supervision solely through pre-training, we leverage the teacher model to adaptively guide exploration during reinforcement learning when naive on-policy rollout fails.
Specifically, we introduce a knowledge-enhanced rollout mechanism that injects teacher-guided trajectories into the on-policy rollout process when naive exploration fails.

% We instantiate two conditional policies from the same underlying model under different rollout contexts. 
We instantiate two conditional policies from the same underlying model under different rollout contexts: a standard rollout policy and a self-enhanced policy.
The standard rollout policy is
\begin{equation}
\pi_S(\cdot \mid x) \triangleq \pi_\theta(\cdot \mid x),
\end{equation}
while the self-enhanced policy is
\begin{equation}
\pi_E(\cdot \mid x,h,y) \triangleq \pi_\theta(\cdot \mid x,h,y).
\end{equation}
Both policies share parameters $\theta$ and differ only in their conditioning context. 
The self-enhanced policy is invoked only in failure-triggered exploration during training.
Concretely, the rollout proceeds as follows.
\begin{enumerate}
    \item \textbf{Adaptive Trigger.}
    For each input $x$, the standard rollout policy $\pi_S$ first performs on-policy rollout by sampling a group of $G$ trajectories:
    \begin{equation}
    \{y_i\}_{i=1}^G \sim \pi_S(\cdot\mid x).
    \end{equation}
    If $\max_{i \in \{1,\dots,G\}} r(y_i)=0$, the rollout is deemed unsuccessful and the exploration mechanism is triggered.

    \item \textbf{Teacher-Guided Hint Generation.}
    Given the input $x$ and the ground-truth answer $y$, the teacher model $\pi_T$ generates a reasoning hint $h$:
    \begin{equation}
        h \sim \pi_T(\cdot \mid x, y).
    \end{equation}

    \item \textbf{Hint-Aware Response Sampling with Rejection.}
    Conditioned on the input $x$, teacher-generated hint $h$, and privileged answer context $y$, 
    % Conditioned on $(x,h,y)$,
    % the input $x$, the teacher-provided hint $h$, and the ground-truth answer $y$, 
    the self-enhanced policy $\pi_E$ generates a proposal trajectory:
    \begin{equation}
        y_h \sim \pi_E(\cdot \mid x, h, y).
    \end{equation}
    % The ground-truth answer is used only as privileged information during failure-triggered exploration in training to anchor teacher-provided hints and recover reward-bearing trajectories when standard on-policy rollouts yield no positive reward. Ground-truth answers are never used at inference time.
    % The ground-truth answer is used only as privileged information in the temporary recovery branch during failure-triggered exploration in training to anchor teacher-provided hints and recover reward-bearing trajectories when standard on-policy rollouts yield no positive reward. It is never used at inference time.
    The ground-truth answer is used only as privileged information in the self-enhanced branch during failure-triggered exploration in training. 
    Its role is to disambiguate and ground teacher-provided hints and facilitate generation of reward-bearing trajectories when standard on-policy rollouts yield no positive reward.
    It is never used at inference time.
    
    The trajectory is accepted if it yields a positive reward; otherwise, sampling is repeated until success or a predefined budget $B$ is reached. 
    Here, $B$ denotes the maximum number of hint-aware resampling attempts in the failure-triggered exploration stage.

    \item \textbf{Trajectory Injection.}
    Accepted hint-aware trajectories are injected into the rollout buffer and jointly optimized with standard on-policy trajectories under the KEPO objective.
\end{enumerate}

An algorithmic description of the adaptive rollout procedure is provided in Appendix~\ref{app:algorithm}.
By increasing the likelihood of encountering reward-bearing trajectories
that are otherwise difficult to obtain for early-stage policies under naive on-policy exploration,
this mechanism helps the model escape the learning cliff 
% while preserving the on-policy nature of reinforcement learning.
while preserving a predominantly on-policy optimization process.
As training progresses and the policy improves, such failure-triggered cases become increasingly rare, implying that the rollout distribution gradually shifts toward fully on-policy data and the algorithm converges to the on-policy learning.

% Taken together, the quality-gated distillation objective and the knowledge-enhanced rollout mechanism form a unified preference optimization framework.
% By jointly addressing optimization instability and exploration failure, KEPO enables effective reinforcement-based post-training even in challenging
% reasoning regimes with insufficient rewards.
% To further contextualize KEPO relative to representative post-training paradigms, a high-level methodological comparison is provided in Appendix~\ref{app:comparison}.
% where sparse rewards alone are insufficient.

Taken together, KEPO forms a unified preference optimization framework that jointly addresses optimization instability and exploration failure, enabling effective reinforcement-based post-training even in challenging reasoning regimes with insufficient rewards.
A high-level methodological comparison with representative post-training paradigms is provided in Appendix~\ref{app:comparison}.

\vspace{-0.4cm}
\section{Experiments}

\vspace{-0.2cm}
\subsection{Datasets and Experimental Setup}
\vspace{-0.2cm}
We conduct experiments on the open-access subset of OmniMedVQA~\cite{hu2024omnimedvqa}, a large-scale medical vision-language benchmark spanning eight imaging modalities. 
To study cross-modality generalization under a controlled and challenging setting, we adopt a single-source training protocol, using MRI as the sole in-domain modality for training. 
At test time, MRI is evaluated as in-domain (ID), while the remaining seven modalities are treated as out-of-distribution (OOD) test sets.
We construct a compact training set of 600 MRI image-question pairs, randomly sampled from the full training split, to study generalization in a low-resource regime.
For evaluation, we sample 300 test pairs per modality, yielding 2,400 test instances in total. 
This MRI-only training and 1 ID vs.\ 7 OOD evaluation protocol provides a systematic stress test for multimodal reasoning transfer under limited supervision.
Additional dataset statistics, category breakdowns, and split construction details are provided in Appendix~\ref{app:data}.

% We conduct experiments on the open-access subset of the OmniMedVQA benchmark~\cite{hu2024omnimedvqa}, a large-scale medical vision-language dataset with 82,059 images and 88,996 VQA pairs spanning eight imaging modalities
% (CT, MRI, X-Ray, Ultrasound, Dermoscopy, Fundus, OCT, and Microscopy)
% and five question categories (Anatomy Identification, Disease Diagnosis, Lesion Grading, Modality Recognition, and Other Biological Attributes).

% To evaluate cross-modality generalization under a controlled and challenging setting,
% we adopt a single-source training protocol, using MRI as the sole in-domain modality.
% At test time, MRI serves as the in-domain (ID) evaluation set, while the remaining seven modalities are treated as out-of-distribution (OOD) evaluation sets.
% As a canonical radiology modality with high structural complexity, MRI provides a non-trivial source domain for transfer to both radiological and non-radiological modalities.
% We construct a compact training set of 600 MRI image-question pairs, randomly sampled from the full training split, to emphasize generalization in a low-resource regime.
% For evaluation, we randomly sample 300 test pairs per modality from the corresponding test split, yielding 2,400 test instances in total.
% This 1 ID vs.\ 7 OOD evaluation protocol provides a systematic stress test for assessing robustness and transferability of multimodal reasoning under limited in-domain supervision.

\vspace{-0.25cm}
\subsection{Implementation Details}
\vspace{-0.25cm}
We use Qwen3-VL-2B as the base vision-language model for all post-training experiments. All compared methods are evaluated under the same low-resource training protocol and shared reward setting to isolate differences in optimization strategies. We adopt a simple sparse rule-based reward with fixed prompting, following prior reasoning-oriented RL work~\cite{shao2024deepseekmath}. Full training hyperparameters, reward designs, and prompt templates are provided in Appendix~\ref{app:train}, Appendix~\ref{app:reward}, and Appendix~\ref{app:prompts}.

\begin{table*}[t]
\centering
\caption{
Performance comparison with baseline methods (accuracy \%) on OmniMedVQA across 8 modalities (MRI, CT, X-Ray, Ultrasound, Dermoscopy, Fundus, OCT, and Microscopy).
All trainable methods are trained solely on MRI (In-Domain), with the remaining 7 modalities serving as Out-of-Distribution (OOD) tests.
All post-training methods use the same backbone (Qwen3-VL-2B) and training protocol, thereby isolating the effect of different optimization strategies.
General-purpose models of different scales are included as capacity references, rather than as post-training baselines.
KEPO achieves the highest Avg. and Avg. (OOD) scores among all compared methods.
Abbreviations: XR: X-Ray, US: Ultrasound, Der: Dermoscopy, Fun: Fundus, Mic: Microscopy.
The best results are highlighted in \textbf{bold}.
}
\label{tab:main_results}
\resizebox{\textwidth}{!}{%
\setlength{\tabcolsep}{3.5pt}
\begin{tabular}{llccc|c|ccccccc|c|c}
\toprule
\multirow{2}{*}{\textbf{Category}} &
\multirow{2}{*}{\textbf{Method}} &
\multirow{2}{*}{\textbf{Size}} &
\multirow{2}{*}{\textbf{Mode}} &
\multirow{2}{*}{\textbf{Teacher Model}} &
\textbf{MRI (ID)} &
\multicolumn{7}{c|}{\textbf{Out-of-Distribution (OOD)}} &
\multirow{2}{*}{\textbf{Avg.}} &
\multirow{2}{*}{\textbf{Avg. (OOD)}} \\
\cmidrule(lr){6-6} \cmidrule(lr){7-13}
& & & & & \textbf{MRI} & \textbf{CT} & \textbf{XR} & \textbf{US} & \textbf{Der} & \textbf{Fun} & \textbf{OCT} & \textbf{Mic} & & \\
\midrule
\multirow{6}{*}{\textbf{General VLM}}
& \multirow{2}{*}{Qwen3-VL~\cite{yang2025qwen3technicalreport}} & \multirow{2}{*}{2B} & Non-thinking & -- & 61.33 & 48.67 & 79.00 & 73.33 & 68.00 & 77.00 & 68.67 & 79.33 & 69.42 & 70.57 \\
&  &  & Thinking & -- & 64.00 & 50.33 & 79.67 & 74.33 & 68.33 & 72.00 & 55.33 & 75.67 & 67.46 & 67.95 \\
\cmidrule(lr){2-5}
& \multirow{2}{*}{Qwen3-VL~\cite{yang2025qwen3technicalreport}} & \multirow{2}{*}{8B} & Non-thinking & -- & 73.67 & 62.00 & 85.67 & 61.67 & 76.67 & 79.00 & 75.33 & 84.00 & 74.75 & 74.91 \\
&  &  & Thinking & -- & 70.00 & 65.67 & 81.33 & 58.67 & 67.33 & 67.00 & 63.33 & 82.00 & 69.42 & 69.33 \\
\cmidrule(lr){2-5}
& \multirow{2}{*}{Qwen3-VL~\cite{yang2025qwen3technicalreport}} & \multirow{2}{*}{32B} & Non-thinking & -- & 79.33 & 68.0 & 86.33 & 66.33 & 83.67 & 84.33 & 85.0 & 81.67 & 79.33 & 79.33 \\
&  &  & Thinking & -- & 77.67 & 66.33 & 84.0 & 62.67 & 77.0 & 76.33 & 78.67 & 80.67 & 75.42 & 75.10 \\
\midrule
\multirow{2}{*}{\textbf{Medical VLM}}
& \multirow{2}{*}{HuatuoGPT~\cite{chen2024huatuogpt}} & \multirow{2}{*}{7B} & Non-thinking & -- & 67.33 & 62.67 & 69.33 & 49.67 & 64.00 & 69.67 & 76.33 & 63.33 & 65.29 & 65.00 \\
& & & Thinking & -- & 64.67 & 62.67 & 71.33 & 45.33 & 53.67 & 63.33 & 72.00 & 63.00 & 62.00 & 61.62 \\
\midrule
\multirow{4}{*}{\textbf{SFT / RL Baseline}}
& SFT~\cite{ouyang2022training} & 2B & Non-thinking & -- & 85.33 & 53.0 & 79.67 & 55.33 & 64.67 & 74.67 & 67.67 & 78.0 & 69.79 & 67.57 \\
\cmidrule(lr){2-5}
& \multirow{2}{*}{GRPO~\cite{shao2024deepseekmath}} & \multirow{2}{*}{2B} & Non-thinking & -- & 80.0 & 48.0 & 78.33 & 72.33 & 71.67 & 81.0 & 73.67 & 80.0 & 73.12 & 72.14 \\
&  &  & Thinking & -- & 79.0 & 51.0 & 78.0 & 73.0 & 75.0 & 81.33 & 73.33 & 84.0 & 74.33 & 73.67 \\
\cmidrule(lr){2-5}
& MM-DAPO~\cite{yu2025dapo} & 2B & Thinking & -- & 64.67 & 46.67 & 78.67 & 76.33 & 70.00 & 72.33 & 58.33 & 74.67 & 67.71 & 68.14 \\
\midrule
\multirow{5}{*}{\textbf{MM-GKD Baseline}}
& Supervised KD ($\lambda=0$)~\cite{agarwal2024gkd} & 2B & Non-thinking & Qwen3-VL-32B & 78.67 & 50.67 & 78.0 & 74.67 & 73.33 & 79.67 & 73.0 & 83.33 & 73.92 & 73.24 \\
\cmidrule(lr){2-5}
& Mixed KD ($\lambda=0.5$)~\cite{agarwal2024gkd} & 2B & Non-thinking & Qwen3-VL-32B & 77.67 & 49.33 & 77.67 & 74.33 & 70.67 & 79.33 & 73.0 & 82.33 & 73.04 & 72.38 \\
\cmidrule(lr){2-5}
& \multirow{2}{*}{On-policy KD ($\lambda=1$)~\cite{agarwal2024gkd}} & \multirow{2}{*}{2B} & Non-thinking & Qwen3-VL-32B & 78.33 & 48.67 & 78.67 & 74.67 & 71.33 & 77.67 & 73.0 & 82.67 & 73.13 & 72.38 \\
& & & Thinking & Qwen3-VL-32B & 59.67 & 52.67 & 79.67 & 75.0 & 67.0 & 61.33 & 58.33 & 73.67 & 65.92 & 66.81 \\
\midrule
\multirow{1}{*}{\textbf{Ours}}
% & \multirow{2}{*}{KEPO - KE ($\tau=0$)} & \multirow{2}{*}{2B} & Non-thinking & Qwen3-VL-32B & 77.67 & 48.33 & 77.0 & 71.67 & 70.0 & 79.0 & 73.0 & 80.33 & 71.33 & 72.13 \\
% & & & Thinking & Qwen3-VL-32B & 81.33 & 53.33 & 81.33 & 72.67 & 82.0 & 74.0 & 75.67 & 84.33 & 75.58 & 74.76 \\
% & KEPO - KE ($\tau=1$) & 2B & Thinking & Qwen3-VL-32B & 83.67 & 55.33 & 81.0 & 72.0 & 74.67 & 82.67 & 77.33 & 83.67 & 76.29 & 75.24 \\
% \cmidrule(lr){2-5}
% (V1, lr=2e-6)
% & \multirow{1}{*}{KEPO ($\tau=0$)} & \multirow{1}{*}{2B} & 
% Thinking & Qwen3-VL-32B & 95.67 & 54.67 & 82.67 & 72.33 & 79.0 & 83.67 & 78.0 & 83.67 & 78.71 & 76.29 \\
& \multirow{1}{*}{\textbf{KEPO ($\tau=1$)}} & \multirow{1}{*}{2B} & 
Thinking & Qwen3-VL-32B & \textbf{96.0} & 62.0 & 85.0 & 74.0 & 78.67 & 82.33 & 82.33 & 81.67 & \textbf{80.25} & \textbf{78.00} \\
% & & & Thinking & Qwen3-VL-32B & 79.67 & 52.67 & 80.0 & 73.33 & 74.67 & 81.0 & 74.0 & 85.0 & 75.04 & 74.38 \\
% & & & Thinking (V1) & Qwen3-VL-32B & 77.0 & 53.0 & 80.33 & 72.67 & 70.33 & 73.33 & 65.0 & 76.0 & 70.96 & 70.09 \\
% & & & Thinking (V1, lr=1e-5) & Qwen3-VL-32B & 96.67 & 69.67 & 41.0 & 42.33 & 62.0 & 69.0 & 78.33 & 59.33 & 64.79 & 60.24 \\
% & & & Thinking (V1, lr=5e-6) & Qwen3-VL-32B & 97.33 & 77.67 & 78.67 & 50.33 & 75.67 & 87.0 & 76.33 & 69.0 & 76.38 & 73.52 \\

\bottomrule
\end{tabular}
}
\vspace{-0.3cm} 
\end{table*}

\vspace{-0.25cm}
\subsection{Baseline Methods and Comparisons}
\vspace{-0.25cm}
We compare against four groups of baselines: (1) general-purpose VLMs from the Qwen3-VL family~\cite{yang2025qwen3technicalreport}, (2) a strong medical-specific VLM, HuatuoGPT-Vision~\cite{chen2024huatuogpt}, (3) standard post-training baselines including SFT~\cite{ouyang2022training}, GRPO~\cite{shao2024deepseekmath}, and MM-DAPO~\cite{yu2025dapo}, and (4) multimodal on-policy distillation baselines based on MM-GKD~\cite{agarwal2024gkd}. Unless otherwise specified, all models are instruction-tuned variants, and we distinguish between thinking and non-thinking configurations based on whether the prompt explicitly elicits intermediate reasoning traces.

MM-DAPO is adapted from the text-only DAPO framework~\cite{yu2025dapo} to the multimodal setting. MM-GKD is adapted from the on-policy distillation framework of Agarwal et al.~\cite{agarwal2024gkd} to the multimodal setting, with $\lambda \in \{0,0.5,1\}$ controlling the balance between supervised and on-policy signals.
For our method, we report both the full KEPO framework and KEPO-KE, which removes the knowledge-enhanced exploration component. Comparing KEPO ($\tau=0$) with KEPO-KE ($\tau=0$) isolates the contribution of knowledge-enhanced exploration, while comparing KEPO-KE ($\tau=1$) with KEPO-KE ($\tau=0$) isolates the effect of quality gating. Ground-truth answers are used only in the failure-triggered exploration branch during training and are never used at inference time.
Detailed model configurations and multimodal adaptations of all baselines are provided in Appendix~\ref{app:baselines}.

\begin{table*}[t]
\centering
\caption{
Ablation study of KEPO on OmniMedVQA (Accuracy \%) under the same MRI-only training protocol.
% The model is trained solely on \textbf{MRI} (In-Domain), with the remaining 7 modalities used for Out-of-Distribution (OOD) evaluation.
% KEPO-KE removes the knowledge-enhanced exploration module, while $\tau$ controls quality-gated distillation.
% Comparing KEPO with KEPO-KE isolates the effect of exploration, and comparing $\tau=1$ with $\tau=0$ evaluates the benefit of quality-aware distillation.
KEPO-KE removes the knowledge-enhanced exploration module, while $\tau$ controls quality-gated distillation. 
Comparing KEPO ($\tau=0$) with KEPO-KE ($\tau=0$) isolates the effect of exploration, while comparing KEPO-KE ($\tau=1$) with KEPO-KE ($\tau=0$) isolates the effect of quality gating.
Abbreviations: XR: X-Ray, US: Ultrasound, Der: Dermoscopy, Fun: Fundus, Mic: Microscopy.
The best results are highlighted in \textbf{bold}.
}
% \caption{
% Ablation study of KEPO on OmniMedVQA (Accuracy \%) under the MRI-only training setting.
% KEPO-KE removes knowledge-enhanced exploration, while $\tau$ controls quality-gated distillation.
% Comparing KEPO with KEPO-KE measures exploration gains, and comparing $\tau=1$ with $\tau=0$ measures the effect of selective distillation.
% Abbreviations: \textbf{XR}: X-Ray, \textbf{US}: Ultrasound, \textbf{Der}: Dermoscopy, \textbf{Fun}: Fundus, \textbf{Mic}: Microscopy.
% }
\label{tab:ablation_results}
\resizebox{\textwidth}{!}{%
\setlength{\tabcolsep}{3.5pt}
\begin{tabular}{llccc|c|ccccccc|c|c}
\toprule
\multirow{2}{*}{\textbf{Category}} &
\multirow{2}{*}{\textbf{Method}} &
\multirow{2}{*}{\textbf{Size}} &
\multirow{2}{*}{\textbf{Mode}} &
\multirow{2}{*}{\textbf{Teacher Model}} &
\textbf{MRI (ID)} &
\multicolumn{7}{c|}{\textbf{Out-of-Distribution (OOD)}} &
\multirow{2}{*}{\textbf{Avg.}} &
\multirow{2}{*}{\textbf{Avg. (OOD)}} \\
\cmidrule(lr){6-6} \cmidrule(lr){7-13}
& & & & & \textbf{MRI} & \textbf{CT} & \textbf{XR} & \textbf{US} & \textbf{Der} & \textbf{Fun} & \textbf{OCT} & \textbf{Mic} & & \\
\midrule
\multirow{5}{*}{\textbf{Configurations}}
& \multirow{2}{*}{KEPO - KE ($\tau=0$)} & \multirow{2}{*}{2B} & Non-thinking & Qwen3-VL-32B & 77.67 & 48.33 & 77.0 & 71.67 & 70.0 & 79.0 & 73.0 & 80.33 & 71.33 & 72.13 \\
& & & Thinking & Qwen3-VL-32B & 81.33 & 53.33 & 81.33 & 72.67 & 82.0 & 74.0 & 75.67 & 84.33 & 75.58 & 74.76 \\
& KEPO - KE ($\tau=1$) & 2B & Thinking & Qwen3-VL-32B & 83.67 & 55.33 & 81.0 & 72.0 & 74.67 & 82.67 & 77.33 & 83.67 & 76.29 & 75.24 \\
\cmidrule(lr){2-5}
% (V1, lr=2e-6)
& \multirow{1}{*}{KEPO ($\tau=0$)} & \multirow{1}{*}{2B} & 
Thinking & Qwen3-VL-32B & 95.67 & 54.67 & 82.67 & 72.33 & 79.0 & 83.67 & 78.0 & 83.67 & 78.71 & 76.29 \\
& \multirow{1}{*}{KEPO ($\tau=1$)} & \multirow{1}{*}{2B} & 
Thinking & Qwen3-VL-32B & \textbf{96.0} & 62.0 & 85.0 & 74.0 & 78.67 & 82.33 & 82.33 & 81.67 & \textbf{80.25} & \textbf{78.00} \\
% & & & Thinking & Qwen3-VL-32B & 79.67 & 52.67 & 80.0 & 73.33 & 74.67 & 81.0 & 74.0 & 85.0 & 75.04 & 74.38 \\
% & & & Thinking (V1) & Qwen3-VL-32B & 77.0 & 53.0 & 80.33 & 72.67 & 70.33 & 73.33 & 65.0 & 76.0 & 70.96 & 70.09 \\
% & & & Thinking (V1, lr=1e-5) & Qwen3-VL-32B & 96.67 & 69.67 & 41.0 & 42.33 & 62.0 & 69.0 & 78.33 & 59.33 & 64.79 & 60.24 \\
% & & & Thinking (V1, lr=5e-6) & Qwen3-VL-32B & 97.33 & 77.67 & 78.67 & 50.33 & 75.67 & 87.0 & 76.33 & 69.0 & 76.38 & 73.52 \\

\bottomrule
\end{tabular}
}
\vspace{-0.3cm} 
\end{table*}

\vspace{-0.3cm} 
\subsection{Results and Discussion}
\vspace{-0.2cm}
\label{sec:results}
Table~\ref{tab:main_results} summarizes the performance of different post-training strategies under both thinking and non-thinking configurations across in-domain (MRI) and out-of-distribution (OOD) modalities.
% Overall, results show that improved performance under distribution shift is not determined by model scale, explicit reasoning prompts, or uniform distillation alone, but depends critically on post-training mechanisms that stabilize reasoning behavior under sparse rewards.
% Overall, the results show that robust multimodal reasoning under distribution shift is not determined by model scale, explicit reasoning prompts, or uniform distillation alone, but depends critically on post-training mechanisms that stabilize exploration and credit assignment under sparse rewards.
Overall, these results show that robust multimodal reasoning under distribution shift is not determined by model scale, explicit reasoning prompts, or uniform distillation alone, but depends critically on post-training mechanisms that 
jointly address exploration and credit assignment under sparse rewards.
% Overall, these results show that robust multimodal reasoning under distribution shift requires post-training mechanisms that jointly address exploration and credit assignment under sparse rewards.

% \vspace{-0.25cm} 
% \paragraph{Thinking Mode Alone Does Not Guarantee Improved Reasoning.}
% Across general-purpose vision-language models, enabling explicit thinking does not consistently improve performance.
% As shown in Table~\ref{tab:main_results}, Qwen3-VL models across different scales exhibit mixed behavior under the thinking configuration, with frequent degradation on OOD modalities.
% While increased model capacity improves overall accuracy, explicit chain-of-thought generation often amplifies existing errors when the underlying policy lacks stable reasoning priors.
% These results suggest that reasoning behavior cannot be reliably induced through prompting alone.

\vspace{-0.25cm}
\paragraph{Thinking Mode Alone Does Not Guarantee Improved Reasoning.}
Across general-purpose vision-language models, enabling thinking does not consistently improve performance. As shown in Table~\ref{tab:main_results}, Qwen3-VL models across different scales exhibit mixed behavior, with frequent degradation on OOD modalities. While increased model capacity improves overall accuracy, explicit chain-of-thought generation can amplify existing errors when the underlying policy lacks stable reasoning priors. This suggests that reasoning behavior cannot be reliably induced through prompting alone.

\vspace{-0.25cm} 
\paragraph{Domain-Specific Supervision Is Insufficient for Robust Reasoning Transfer.}
Despite being explicitly trained for medical reasoning, HuatuoGPT-Vision fails to demonstrate robust cross-modality generalization under the single-source training protocol.
This result suggests that large-scale domain-specific supervision, in the absence of post-training mechanisms for reasoning stabilization, is insufficient to support transferable reasoning under distribution shift.

% \vspace{-0.25cm} 
% \paragraph{Behavior of Uniform On-Policy Distillation under Explicit Reasoning.}
% Uniform on-policy distillation exhibits a pronounced discrepancy between non-thinking and thinking configurations.
% While MM-GKD achieves competitive performance without explicit reasoning, its performance degrades substantially once thinking is enabled; for example, pure on-policy distillation ($\lambda = 1$) shows a marked drop in average accuracy.
% This indicates that uniformly distilling all student-generated trajectories is ill-suited for reasoning-intensive settings.
% When early rollouts contain logical or perceptual errors, distilling teacher outputs conditioned on flawed intermediate contexts introduces noisy and misaligned gradients, hindering the stable emergence of coherent chain-of-thought reasoning.

\vspace{-0.25cm}
\paragraph{Behavior of Uniform On-Policy Distillation under Explicit Reasoning.}
Uniform on-policy distillation exhibits a clear discrepancy between non-thinking and thinking configurations. While MM-GKD achieves competitive performance without explicit reasoning, its performance degrades substantially once thinking is enabled; for example, pure on-policy distillation ($\lambda = 1$) shows a marked drop in average accuracy. This indicates that uniformly distilling all student-generated trajectories is ill-suited for reasoning-intensive settings. When early rollouts contain logical or perceptual errors, teacher supervision conditioned on flawed intermediate contexts can introduce noisy and misaligned gradients, hindering the stable emergence of coherent chain-of-thought reasoning.

\begin{figure*}[t]
    \centering
    \begin{subfigure}[t]{0.48\textwidth}
        \centering
        \includegraphics[width=\textwidth]{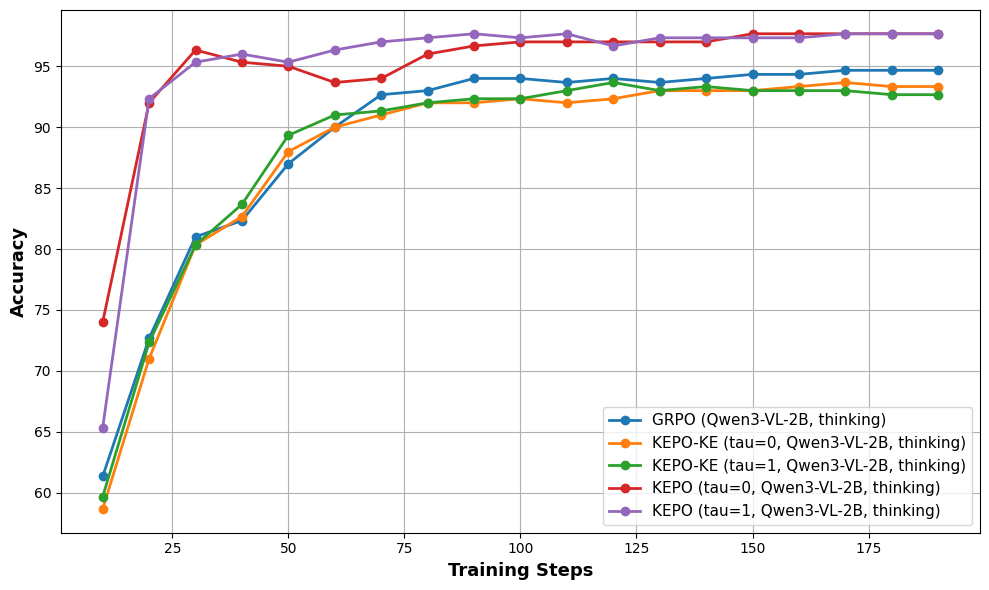}
        \caption{Accuracy on the in-domain MRI modality.}
        \label{fig:mri_accuracy}        
    \end{subfigure}
    \hfill
    \begin{subfigure}[t]{0.48\textwidth}
        \centering
        \includegraphics[width=\textwidth]{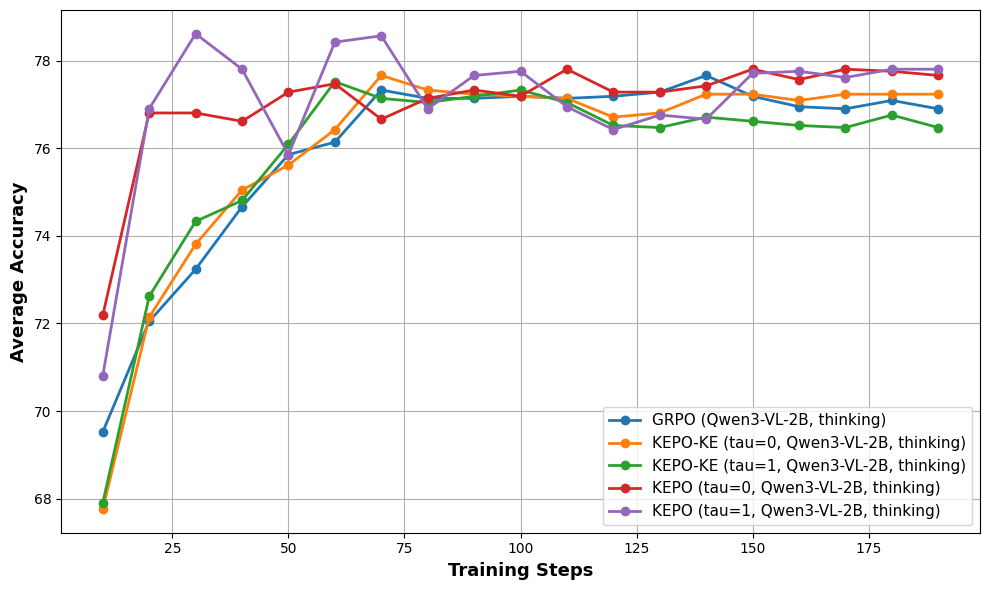}
        \caption{Average accuracy across out-of-distribution modalities (averaged over 7 modalities).}
        \label{fig:ood_accuracy}
    \end{subfigure}
    \vspace{-0.1cm}
    \caption{
    Training dynamics under different post-training strategies. All experiments are trained for 5 epochs, with checkpoints evaluated every 10 steps on two metrics: Left: accuracy on the in-domain (ID) MRI modality. Right: average accuracy across out-of-distribution (OOD) modalities.
    Compared to GRPO, KEPO-KE achieves stronger ID and OOD performance in the early training stage, while the full KEPO framework consistently outperforms all baselines throughout training and converges faster.
    Performance gains largely stabilize within the current training horizon.
% These observations point to complementary effects of credit assignment
% and exploration on reasoning behavior under distribution shift.
    }
    \label{fig:training_dynamics}
    \vspace{-0.6cm}
\end{figure*}

\vspace{-0.25cm}
\paragraph{KEPO Enables Stable and Transferable Reasoning Emergence.}
KEPO consistently benefits from the thinking configuration and achieves the strongest overall performance across OOD modalities. Under the same low-resource setting, KEPO outperforms all baselines in both average accuracy and average OOD accuracy. It also exhibits improved training stability and robustness to distribution shift (Figure~\ref{fig:training_dynamics}), reflecting the complementary effects of knowledge-enhanced exploration and quality-gated distillation. Together, these components enable reasoning behaviors that generalize across modalities.

% \vspace{-0.25cm} 
% \paragraph{KEPO Enables Stable and Transferable Reasoning Emergence.}
% In contrast, KEPO consistently benefits from the thinking configuration and achieves the strongest overall performance across OOD modalities.
% Under the same low-resource training regime, KEPO with thinking outperforms all baselines in both average accuracy and average OOD accuracy.
% Beyond final performance, KEPO also exhibits improved training stability and robustness to distribution shift (Figure~\ref{fig:training_dynamics}).
% This advantage arises from the complementary roles of its two components:
% the knowledge-enhanced exploration mechanism increases exposure to reward-bearing trajectories that are rarely discovered through naive on-policy rollout,
% while quality-gated distillation stabilizes learning by providing dense, reward-aligned credit assignment on those trajectories.
% Together, these mechanisms enable the emergence of coherent reasoning behaviors that generalize across modalities.

% Overall, the results demonstrate that robust reasoning under distribution shift does not emerge from explicit prompting, uniform distillation, or domain-specific supervision alone,
% but requires post-training mechanisms that jointly address exploration and credit assignment under sparse rewards, as instantiated by KEPO.

\vspace{-0.3cm}
\subsection{Ablation Study: How Reasoning Emerges during Post-training}
\vspace{-0.2cm}
We conduct component-wise ablations to disentangle the roles of quality-gated distillation and knowledge-enhanced exploration in KEPO. 
We use GRPO (thinking) as a sparse-reward reasoning baseline, representing reasoning behaviors that emerge purely from trajectory-level rewards without teacher guidance.
As summarized in Table~\ref{tab:ablation_results}, KEPO-KE removes the knowledge-enhanced exploration module, while varying $\tau \in \{0,1\}$ controls the strength of quality gating. 
Comparing KEPO ($\tau=0$) with KEPO-KE ($\tau=0$) isolates the contribution of knowledge-enhanced exploration, while comparing KEPO-KE ($\tau=1$) with KEPO-KE ($\tau=0$) isolates the effect of quality gating.
We further compare training dynamics in Figure~\ref{fig:training_dynamics} to understand how these components affect optimization stability and reasoning emergence.
Further discussion of the ablation setup and motivation is provided in Appendix~\ref{app:ablation_setup}.

\vspace{-0.25cm}
\paragraph{Quantitative Component Effects.}
Table~\ref{tab:ablation_results} shows that both quality-gated distillation and knowledge-enhanced exploration contribute substantially to KEPO. Under fixed $\tau=0$, adding knowledge-enhanced exploration improves Avg. OOD accuracy from 74.76 to 76.29, confirming the benefit of guided exploration. Under KEPO-KE, increasing the gating threshold from $\tau=0$ to $\tau=1$ improves Avg. OOD accuracy from 74.76 to 75.24, indicating that stricter quality gating yields more robust selective supervision.

\vspace{-0.25cm}
\paragraph{Training Dynamics and Stability.}
Figure~\ref{fig:training_dynamics} shows that KEPO improves both optimization stability and out-of-distribution generalization throughout training, while KEPO-KE already provides clear gains over GRPO in the early stage.
On the in-domain MRI task, GRPO and KEPO variants all converge rapidly to high accuracy, indicating that the source modality provides sufficiently dense learning signals for effective optimization. Both KEPO-KE and KEPO reach strong in-domain performance early and maintain comparable final accuracy, suggesting that teacher-guided mechanisms do not harm source-domain learning.

The differences become more pronounced on OOD modalities. GRPO improves more slowly and saturates earlier, highlighting the difficulty of learning transferable reasoning patterns from sparse trajectory-level rewards alone. Relative to GRPO, KEPO-KE achieves faster early-stage improvement, indicating that quality-gated distillation provides additional reward-aligned learning signals beyond sparse rewards. The full KEPO framework further achieves higher OOD accuracy and reduced late-stage degradation, showing that knowledge-enhanced exploration improves robustness under distribution shift by increasing exposure to reward-bearing trajectories during training. Across both KEPO-KE and KEPO, varying $\tau$ reveals a trade-off between early gains and long-term stability: less restrictive gating can lead to higher early peaks, while stricter gating produces smoother optimization and more stable late-stage performance.
Overall, these dynamics suggest that quality-gated distillation improves credit assignment, while knowledge-enhanced exploration further strengthens generalization by alleviating exploration failure under sparse rewards.

\vspace{-0.4cm}
\section{Conclusion}

\enlargethispage{\baselineskip}
\vspace{-0.3cm} 
We propose Knowledge-Enhanced Preference Optimization (KEPO), a reinforcement-based post-training framework designed to address two persistent bottlenecks in sparse-reward reasoning: unstable optimization from trajectory-level supervision and exploration collapse in complex reasoning regimes.
% KEPO is treated as an implicit token-level reward shaping approach via exploration-time conditioning. 
KEPO unifies quality-gated on-policy distillation with knowledge-enhanced exploration, allowing dense teacher supervision to reinforce only reward-aligned behaviors while actively steering exploration toward informative trajectories that early-stage policies rarely encounter.
This design enables stable learning progress even when naive on-policy reinforcement learning stagnates under sparse rewards.
Empirical results on challenging medical vision-language reasoning benchmarks demonstrate that KEPO consistently outperforms strong reinforcement learning and on-policy distillation baselines.
More broadly, KEPO offers a general recipe for injecting structured knowledge into reinforcement-based post-training without sacrificing the on-policy nature of learning.

% We propose Knowledge-Enhanced Preference Optimization (KEPO), a reinforcement-based post-training framework that addresses two central challenges in sparse-reward reasoning: unstable optimization under coarse trajectory-level supervision and exploration failure in complex reasoning regimes.
% KEPO integrates quality-gated on-policy distillation with knowledge-enhanced exploration within a unified group-based policy optimization framework.
% By selectively applying dense teacher guidance only to reward-aligned trajectories and actively increasing exposure to reward-bearing trajectories rarely discovered by naive on-policy exploration, KEPO improves credit assignment while avoiding contaminated reasoning contexts.
% Together, these components enable stable and effective learning in settings where
% sparse rewards alone lead to training stagnation.
% Experiments on challenging medical vision-language reasoning benchmarks demonstrate consistent improvements over strong reinforcement learning and on-policy distillation baselines.
% Beyond the medical domain, KEPO provides a general strategy for integrating structured knowledge into reinforcement-based post-training without compromising on-policy exploration.

\newpage
\bibliography{refs}

@misc{zheng2026scopesignalcalibratedonpolicydistillation,
      title={SCOPE: Signal-Calibrated On-Policy Distillation Enhancement with Dual-Path Adaptive Weighting}, 
      author={Binbin Zheng and Xing Ma and Yiheng Liang and others},
      year={2026},
      eprint={2604.10688},
      archivePrefix={arXiv},
      primaryClass={cs.LG},
      url={https://arxiv.org/abs/2604.10688}, 
}

@article{Guo_2025,
   title={DeepSeek-R1 incentivizes reasoning in LLMs through reinforcement learning},
   volume={645},
   ISSN={1476-4687},
   url={http://dx.doi.org/10.1038/s41586-025-09422-z},
   DOI={10.1038/s41586-025-09422-z},
   number={8081},
   journal={Nature},
   publisher={Springer Science and Business Media LLC},
   author={Guo, Daya and Yang, Dejian and Zhang, Haowei and others},
   year={2025},
   month=Sept, pages={633–638} }

@misc{yang2025qwen3technicalreport,
      title={Qwen3 Technical Report}, 
      author={An Yang and Anfeng Li and Baosong Yang and others},
      year={2025},
      eprint={2505.09388},
      archivePrefix={arXiv},
      primaryClass={cs.CL},
      url={https://arxiv.org/abs/2505.09388}, 
}

@article{yu2025dapo,
  title={Dapo: An open-source llm reinforcement learning system at scale},
  author={Yu, Qiying and Zhang, Zheng and Zhu, Ruofei and others},
  journal={arXiv preprint arXiv:2503.14476},
  year={2025}
}

@article{ma2025interleaved,
  title={Learning What Reinforcement Learning Can't: Interleaved Online Fine-Tuning for Hardest Questions},
  author={Ma, Lu and Liang, Hao and Qiang, Meiyi and others},
  journal={arXiv preprint arXiv:2506.07527},
  year={2025},
}

@article{liu2025uft,
  title={UFT: Unifying Supervised and Reinforcement Fine-Tuning},
  author={Liu, Mingyang and Farina, Gabriele and Ozdaglar, Asuman},
  journal={arXiv preprint arXiv:2505.16984},
  year={2025},
}

@article{yan2025learning,
  title={Learning to Reason Under Off-Policy Guidance},
  author={Yan, Jianhao and Li, Yafu and Hu, Zican and others},
  journal={arXiv preprint arXiv:2504.14945},
  year={2025},
}

@article{chen2025stepwise,
  title={Step-wise Adaptive Integration of Supervised Fine-Tuning and Reinforcement Learning for Task-Specific LLMs},
  author={Chen, Jack and Liu, Fazhong and Liu, Naruto and others},
  journal={arXiv preprint arXiv:2505.13026},
  year={2025},
}

@article{chen2025sftrl,
  title={SFT or RL? An Early Investigation into Training R1-Like Reasoning Large Vision-Language Models},
  author={Chen, Hardy and Tu, Haoqin and others},
  journal={arXiv preprint arXiv:2504.11468},
  year={2025},
}

@article{cai2025backtracking,
  title={How Much Backtracking is Enough? Exploring the Interplay of SFT and RL in Enhancing LLM Reasoning},
  author={Cai, Hongyi James and Wang, Junlin and Chen, Xiaoyin and Dhingra, Bhuwan},
  journal={arXiv preprint arXiv:2505.24273},
  year={2025},
}

@article{huang2025blending,
  title={Blending Supervised and Reinforcement Fine-Tuning with Prefix Sampling},
  author={Huang, Zeyu and Cheng, Tianhao and Qiu, Zihan and others},
  journal={arXiv preprint arXiv:2507.01679},
  year={2025},
}

@article{zhang2025scafgrpo,
  title     = {Scaf-GRPO: Scaffolded Group Relative Policy Optimization for Enhancing LLM Reasoning},
  author    = {Zichen Zhang and Sitong Wu and Yinghao Zhu and others},
  journal   = {arXiv preprint arXiv:2510.19807},
  year      = {2025}
}

@article{zhang2025stephint,
  title     = {StepHint: Multi-level Stepwise Hints Enhance Reinforcement Learning to Reason},
  author    = {Kaiyi Zhang and Ang Lv and Jinpeng Li and others},
  journal   = {arXiv preprint arXiv:2507.02841},
  year      = {2025}
}

@article{xuechen2025bread,
  title     = {BREAD: Branched Rollouts from Expert Anchors Bridge SFT \& RL for Reasoning},
  author    = {Xuechen Zhang and Zijian Huang and Yingcong Li and others},
  journal   = {arXiv preprint arXiv:2506.17211},
  year      = {2025}
}

@inproceedings{ahmadian2024back,
  title     = {Back to Basics: Revisiting REINFORCE-Style Optimization for Learning from Human Feedback in LLMs},
  author    = {Ahmadian, Arash and Cremer, Chris and Gallé, Matthias and others},
  booktitle = {Proceedings of the 62nd Annual Meeting of the Association for Computational Linguistics (ACL)},
  year      = {2024}
}

@article{ouyang2022training,
  title   = {Training Language Models to Follow Instructions with Human Feedback},
  author  = {Ouyang, Long and Wu, Jeffrey and Jiang, Xu and others},
  journal = {Advances in Neural Information Processing Systems},
  volume  = {35},
  year    = {2022}
}

@article{shao2024deepseekmath,
  title   = {DeepSeekMath: Pushing the Limits of Mathematical Reasoning in Open Language Models},
  author  = {Shao, Zhihong and Wang, Peiyi and Zhu, Qihao and others},
  journal = {arXiv preprint arXiv:2402.03300},
  year    = {2024}
}

@article{hinton2015distilling,
  title   = {Distilling the Knowledge in a Neural Network},
  author  = {Hinton, Geoffrey and Vinyals, Oriol and Dean, Jeff},
  journal = {arXiv preprint arXiv:1503.02531},
  year    = {2015}
}

@inproceedings{agarwal2024gkd,
  title     = {On-Policy Distillation of Language Models: Learning from Self-Generated Mistakes},
  author    = {Agarwal, Rishabh and Vieillard, Nino and Zhou, Yongchao and others},
  booktitle = {International Conference on Learning Representations (ICLR)},
  year      = {2024}
}

@misc{thinkingmachines2025,
  title        = {On-Policy Distillation},
  author       = {{Thinking Machines}},
  year         = {2025},
  howpublished = {\url{https://thinkingmachines.ai/blog/on-policy-distillation/}}
}

@misc{huggingface2025,
  title        = {On-Policy Distillation Demo},
  author       = {{HuggingFace H4 Team}},
  year         = {2025},
  howpublished = {\url{https://huggingface.co/spaces/HuggingFaceH4/on-policy-distillation}}
}

@article{lightman2023let,
  title   = {Let’s Verify Step by Step},
  author  = {Lightman, Hunter and Kosaraju, Vineet and Burda, Yura and others},
  journal = {arXiv preprint arXiv:2305.20050},
  year    = {2023}
}

@article{lai2025medr1,
  title   = {Med-R1: Reinforcement Learning for Generalizable Medical Reasoning in Vision-Language Models},
  author  = {Lai, Yuxiang and Zhong, Jike and others},
  journal = {arXiv preprint arXiv:2503.13939},
  year    = {2025}
}

@article{pan2025medvlmr1,
  title   = {MedVLM-R1: Incentivizing Medical Reasoning Capability of Vision-Language Models (VLMs) via Reinforcement Learning},
  author  = {Pan, Jiazhen and Liu, Che and Wu, Junde and others},
  journal = {arXiv preprint arXiv:2502.19634},
  year    = {2025}
}

@article{bousselham2025vold,
  title   = {VOLD: Reasoning Transfer from LLMs to Vision-Language Models via On-Policy Distillation},
  author  = {Bousselham, Walid and Kuehne, Hilde and Schmid, Cordelia},
  journal = {arXiv preprint arXiv:2510.23497},
  year    = {2025}
}

@inproceedings{hu2024omnimedvqa,
  title     = {OmniMedVQA: A New Large-Scale Comprehensive Evaluation Benchmark for Medical LVLM},
  author    = {Hu, Yutao and Li, Tianbin and Lu, Quanfeng and others},
  booktitle = {Proceedings of the IEEE/CVF Conference on Computer Vision and Pattern Recognition (CVPR)},
  year      = {2024}
}

@inproceedings{chen2024huatuogpt,
    title = "Towards Injecting Medical Visual Knowledge into Multimodal {LLM}s at Scale",
    author = "Chen, Junying  and
      Gui, Chi  and
      Ouyang, Ruyi  and
      others",
    booktitle = "Proceedings of the 2024 Conference on Empirical Methods in Natural Language Processing",
    month = nov,
    year = "2024",
    publisher = "Association for Computational Linguistics",
    url = "https://aclanthology.org/2024.emnlp-main.418/",
    doi = "10.18653/v1/2024.emnlp-main.418",
    pages = "7346--7370",
}

@inproceedings{liu2024improved,
  title     = {Improved Baselines with Visual Instruction Tuning},
  author    = {Liu, Haotian and Li, Chunyuan and Li, Yuheng and Lee, Yong Jae},
  booktitle = {Proceedings of the IEEE/CVF Conference on Computer Vision and Pattern Recognition (CVPR)},
  pages     = {26296--26306},
  year      = {2024}
}
\bibliographystyle{unsrtnat}

% \section*{References}
% References follow the acknowledgments in the camera-ready paper. Use unnumbered first-level heading for
% the references. Any choice of citation style is acceptable as long as you are
% consistent. It is permissible to reduce the font size to \verb+small+ (9 point)
% when listing the references.
% Note that the Reference section does not count towards the page limit.
% \medskip
% {
% \small

% [1] Alexander, J.A.\ \& Mozer, M.C.\ (1995) Template-based algorithms for
% connectionist rule extraction. In G.\ Tesauro, D.S.\ Touretzky and T.K.\ Leen
% (eds.), {\it Advances in Neural Information Processing Systems 7},
% pp.\ 609--616. Cambridge, MA: MIT Press.

% [2] Bower, J.M.\ \& Beeman, D.\ (1995) {\it The Book of GENESIS: Exploring
%   Realistic Neural Models with the GEneral NEural SImulation System.}  New York:
% TELOS/Springer--Verlag.

% [3] Hasselmo, M.E., Schnell, E.\ \& Barkai, E.\ (1995) Dynamics of learning and
% recall at excitatory recurrent synapses and cholinergic modulation in rat
% hippocampal region CA3. {\it Journal of Neuroscience} {\bf 15}(7):5249-5262.
% }

%%%%%%%%%%%%%%%%%%%%%%%%%%%%%%%%%%%%%%%%%%%%%%%%%%%%%%%%%%%%
\newpage
\appendix
\onecolumn

\appendix

\section{Algorithmic Description of Knowledge-Enhanced Rollout}
\label{app:algorithm}
Algorithm~\ref{alg:kepo_rollout} provides an illustrative algorithmic summary of the adaptive knowledge-enhanced rollout procedure described in Section~\ref{sec:kepo-exploration}.

\begin{algorithm}[]
\caption{Adaptive Knowledge-Enhanced Rollout in KEPO}
\label{alg:kepo_rollout}
\begin{algorithmic}[1]
\STATE \textbf{Input:} input $x$, ground-truth answer $y$, student policy $\pi_\theta$, teacher policy $\pi_T$, group size $G$, rejection budget $B$
\STATE \textbf{Output:} rollout buffer $\mathcal{D}$

\STATE Sample $\{y_i\}_{i=1}^G \sim \pi_\theta(\cdot \mid x)$
\IF{$\max_{i} r(y_i) > 0$}
    \STATE Add $\{(x, y_i)\}_{i=1}^G$ to $\mathcal{D}$
\ELSE
    \STATE Sample hint $h \sim \pi_T(\cdot \mid x, y)$
    \FOR{$b = 1$ {\bfseries to} $B$}
        \STATE Sample $y_h \sim \pi_\theta(\cdot \mid x, h, y)$
        \IF{$r(y_h) > 0$}
            \STATE Add $(x, y_h)$ to $\mathcal{D}$
            \STATE \textbf{break}
        \ENDIF
    \ENDFOR
    % \STATE Add subset of $\{(x, y_i)\}_{i=1}^G$ to $\mathcal{D}$ to ensure a total of $G$ responses
    % \STATE \textit{// If no reward-positive sample is found within budget $B$, no trajectory is added}
    \STATE Fill the remaining slots in $\mathcal{D}$ with samples from $\{(x, y_i)\}_{i=1}^G$ until $|\mathcal{D}|=G$
    \STATE \textit{// If no reward-positive sample is found within budget $B$, no hint-aware trajectory is added}
\ENDIF
\end{algorithmic}
\end{algorithm}

\section{Compute Resources}
\label{app:compute}
All experiments are conducted on an internal on-premise GPU server equipped with NVIDIA A100-SXM4-80GB GPUs (80GB memory per GPU). 
Training is performed using multi-GPU setups when applicable. 
Each training run typically takes around 4 hours, depending on the method and configuration. 
No external cloud resources are used.

\section{Broader Impact}
\label{app:broader}
This work aims to improve multimodal reasoning capabilities in vision-language models, which may benefit applications such as medical question answering and clinical decision support. 

However, the deployment of such models may also introduce risks, particularly in medical contexts where incorrect predictions could lead to misleading conclusions or unsafe decisions. 

We emphasize that these models should not be used as standalone decision-making systems and should be applied with appropriate human oversight.

\section{Limitations}
\label{app:limitations}
This work has several limitations. First, the effectiveness of the method may depend on the quality of teacher-generated hints, which can vary across settings. 
Second, while the method demonstrates strong performance on the evaluated benchmark, additional validation on more datasets could further confirm its generality. 
Third, the rejection sampling procedure introduces some additional computational overhead compared to standard rollout-based approaches, though this overhead is relatively modest in practice.

\section{Method Comparisons}
\label{app:comparison}

Table~\ref{tab:paradigm_compare} provides a high-level comparison between KEPO and representative post-training paradigms for multimodal reasoning. The comparison highlights key differences in optimization data sources, supervision signals, and exploration mechanisms.

\begin{table}[H]
\centering
\small
\setlength{\tabcolsep}{5pt}
\caption{
High-level comparison of representative post-training paradigms for multimodal reasoning.
KEPO combines on-policy reinforcement learning, quality-gated dense supervision, and adaptive exploration recovery within a unified framework.
}
\label{tab:paradigm_compare}
\begin{tabular}{lcccc}
\toprule
\textbf{Property} & \textbf{SFT} & \textbf{GRPO} & \textbf{MM-GKD} & \textbf{KEPO} \\
\midrule
On-policy Optimization         & \ding{55} & \ding{51} & \ding{51} & \ding{51} \\
Reward-driven Learning         & \ding{55} & \ding{51} & \ding{55} & \ding{51} \\
Teacher-guided Dense Supervision      & \ding{55} & \ding{55} & \ding{51} & \ding{51} \\
Quality-aware Distillation     & \ding{55} & \ding{55} & \ding{55} & \ding{51} \\
Adaptive Exploration Recovery  & \ding{55} & \ding{55} & \ding{55} & \ding{51} \\
% External Teacher Required      & \ding{55} & \ding{55} & \ding{51} & \ding{51} \\
\bottomrule
\end{tabular}
\end{table}

\section{Additional Experimental Details}

\subsection{Dataset Details}
\label{app:data}
We conduct experiments on the open-access subset of the OmniMedVQA benchmark~\cite{hu2024omnimedvqa}, a large-scale medical vision-language dataset with 82,059 images and 88,996 VQA pairs spanning eight imaging modalities (CT, MRI, X-Ray, Ultrasound, Dermoscopy, Fundus, OCT, and Microscopy) and five question categories (Anatomy Identification, Disease Diagnosis, Lesion Grading, Modality Recognition, and Other Biological Attributes).

To evaluate cross-modality generalization under a controlled and challenging setting, we adopt a single-source training protocol, using MRI as the sole training modality.
As a canonical radiology modality with high structural complexity, MRI provides a non-trivial source domain for transfer to both radiological and non-radiological modalities. 
At test time, MRI serves as the in-domain evaluation set, while the remaining seven modalities are treated as OOD test sets.

We construct a compact training set of 600 MRI image-question pairs, randomly sampled from the full training split, to study generalization in a low-resource regime.
% For evaluation, we randomly sample 300 test pairs per modality from the corresponding test split, yielding 2,400 test instances in total.
For evaluation, we randomly sample 300 test pairs from each modality-specific test split, resulting in 2,400 total test instances.
This 1-ID vs.\ 7-OOD evaluation protocol provides a systematic stress test for assessing robustness and transferability of multimodal reasoning under limited in-domain supervision.

\subsection{Training Details}
\label{app:train}
We use Qwen3-VL-2B as the base vision–language model. For all reinforcement learning experiments, we sample $G=8$ trajectories per input with a maximum generation length of 1024 tokens and a per-device batch size of 2. We set the learning rate to 2e-6.
All post-training experiments reported in Table~\ref{tab:main_results} are conducted in a deliberately low-resource setting. Specifically, we train for a single epoch on 600 training samples using 4 GPUs. For the ablation studies in Figures~\ref{fig:mri_accuracy} and~\ref{fig:ood_accuracy}, we train on the same data for 5 epochs using 8 GPUs.
Although the post-training horizon is short, this is not restrictive for the current benchmark, where the reinforcement learning rewards on both in-domain and out-of-distribution tasks are observed to saturate within this regime.

\subsection{Reward Design}
\label{app:reward}
We employ a simple rule-based reward composed of format and accuracy components.
Specifically, the model is prompted to produce an explicit reasoning trace and a final answer using a fixed structured format.
A binary format reward verifies the presence of the required reasoning and answer fields,
while a binary accuracy reward checks whether the predicted answer matches the ground-truth option.
The total reward is the sum of these components and remains sparse and outcome-based,
without providing intermediate supervision over reasoning steps.
This design ensures that all methods operate under comparable reward signals,
allowing us to isolate the effect of post-training strategies rather than reward engineering.

\subsection{Baseline Details}
\label{app:baselines}
We compare against the following baselines in detail:

\begin{itemize}
    \item \textbf{General-Purpose VLMs.} We include models from the Qwen3-VL family~\cite{yang2025qwen3technicalreport} to assess scaling behavior and general reasoning capability. We consider Qwen3-VL-2B, Qwen3-VL-8B, and Qwen3-VL-32B, each evaluated under standard and thinking decoding.

    \item \textbf{Medical-Specific VLMs.} We compare against HuatuoGPT-Vision~\cite{chen2024huatuogpt}, a strong medical vision-language model built upon Qwen2.5-VL-7B and explicitly optimized for complex medical reasoning tasks with large-scale medical supervision.

    \item \textbf{Fine-Tuning Baselines.} We fine-tune the Qwen3-VL-2B instruction model using standard Supervised Fine-Tuning (SFT)~\cite{ouyang2022training} and Group Relative Policy Optimization (GRPO)~\cite{shao2024deepseekmath}. All fine-tuning methods use the identical training set of 600 MRI image-question pairs.

    \item \textbf{MM-DAPO.} We evaluate Multi-Modal DAPO (MM-DAPO), adapted from the text-only DAPO framework~\cite{yu2025dapo} to the multimodal setting. MM-DAPO serves as an exploration-oriented reinforcement learning baseline under sparse-reward supervision.

    \item \textbf{MM-GKD Variants.} We evaluate Multi-Modal Generalized Knowledge Distillation (MM-GKD), adapted from the on-policy distillation framework of Agarwal et al.~\cite{agarwal2024gkd} to the multimodal setting. We vary the interpolation coefficient $\lambda \in \{0,0.5,1\}$, corresponding to supervised, mixed, and on-policy settings.
    For the applicable variant (thinking mode is only applied in on-policy knowledge distillation due to the lack of off-policy data with explicit reasoning traces), we evaluate both thinking and non-thinking decoding modes.

    \item \textbf{KEPO Variants.} We evaluate the full KEPO framework as well as KEPO-KE, which removes knowledge-enhanced exploration while retaining the remaining optimization components, under both thinking and non-thinking configurations.
\end{itemize}

\subsection{Additional Ablation Motivation}
\label{app:ablation_setup}

The base model used in our experiments is instruction-tuned and does not explicitly optimize for chain-of-thought reasoning. Under this setting, simply enabling the thinking mode does not consistently improve performance when trained with supervised or uniform distillation objectives, suggesting that reasoning behaviors do not reliably emerge from prompting alone.

Reinforcement learning methods such as GRPO can nevertheless induce effective reasoning patterns through reward-driven credit assignment. We therefore adopt GRPO with thinking as a reference baseline that captures reasoning emerging purely from sparse, trajectory-level rewards.

Building on this reinforcement-based baseline, our ablation study examines how quality-gated supervision and knowledge-enhanced exploration influence training dynamics and generalization. All variants use the same training data, backbone architecture, and decoding strategy, allowing controlled comparison across components.

\section{Prompt Templates}
\label{app:prompts}

This appendix lists the prompt templates used in our experiments.
All prompts are fixed across methods unless otherwise specified.

\begin{promptbox}[promptblue]{VQA Prompt with Explicit Reasoning (Thinking Mode)}
Your task:
1. Think through the question step by step, enclose your reasoning process
   in <think>...</think> tags.
2. Then provide the correct single-letter choice (A, B, C, D, ...)
   inside <answer>...</answer> tags.
3. No extra information or text outside of these tags.
\end{promptbox}

\vspace{0.5em}

\begin{promptbox}[promptgray]{Standard VQA Prompt (Non-thinking)}
Your task:
1. Provide the correct single-letter choice (A, B, C, D, ...)
   inside <answer>...</answer> tags.
2. No extra information or text outside of this tag.
\end{promptbox}

\vspace{0.5em}

\begin{promptbox}[promptgreen]{Teacher Hint Generation Prompt}
Your task:
1. You will get a correct answer for the question.
2. Please provide a hint for the question based on the correct answer,
   inside <hint>...</hint> tags.
3. No extra information or text outside of these tags.

The ground truth answer is {answer}.
\end{promptbox}

\vspace{0.5em}

% \newpage
% \clearpage
\begin{promptbox}[promptorange]{Hint-Aware VQA Prompt with Reasoning}
Your task:
1. Read the hint provided in <hint>...</hint> tags and the ground truth
   answer provided in <answer>...</answer> tags.
2. Think through the question step by step but do not explicitly mention
   the hint or the ground truth answer, enclose your reasoning process
   in <think>...</think> tags.
3. Then provide the correct single-letter choice (A, B, C, D, ...)
   inside <answer>...</answer> tags.
4. No extra information or text outside of these tags.

The hint is <hint>{hint}</hint> and the ground truth answer is
<answer>{answer}</answer>.
\end{promptbox}

% \section{Technical appendices and supplementary material}
% Technical appendices with additional results, figures, graphs, and proofs may be submitted with the paper submission before the full submission deadline (see above). You can upload a ZIP file for videos or code, but do not upload a separate PDF file for the appendix. There is no page limit for the technical appendices. 

% Note: Think of the appendix as ``optional reading'' for reviewers. The paper must be able to stand alone without the appendix; for example, adding critical experiments that support the main claims to an appendix is inappropriate. 

%%%%%%%%%%%%%%%%%%%%%%%%%%%%%%%%%%%%%%%%%%%%%%%%%%%%%%%%%%%%

% \newpage
% \input{checklist.tex}

\end{document}